\documentclass[runningheads]{llncs}

\usepackage{accv}

\usepackage{accvabbrv}

\usepackage{graphicx}
\usepackage{booktabs}

\usepackage[accsupp]{axessibility}  
\usepackage{booktabs}
\usepackage{multirow}

\usepackage{hyperref}

\usepackage{orcidlink}

\begin{document}

\title{SARFusion: Scene-Aware Routing Fusion for Robust Camera-LiDAR 3D Object Detection} 

\titlerunning{SARFusion: Scene-Aware Routing Fusion}

\author{Yuting Zhao\inst{1,2} \and
Ziyi Zheng\inst{3} \and
Shuxiao Li\inst{1}\thanks{Corresponding author: shuxiao.li@ia.ac.cn}}

\authorrunning{Y. Zhao et al.}

\institute{Institute of Automation, Chinese Academy of Sciences, Beijing, China \and
School of Artificial Intelligence, University of Chinese Academy of Sciences, Beijing, China \and
Wuhan College, Wuhan, China\\
\email{\{zhaoyuting2023,shuxiao.li\}@ia.ac.cn, 542197655@qq.com}}

\maketitle

\begin{abstract}

Camera-LiDAR fusion has become a prevailing paradigm for 3D object detection in autonomous driving. However, existing fusion detectors often establish strong inter-modality dependencies by decoding object queries from tightly coupled multimodal representations. Under corrupted driving conditions, such dependencies make the detector vulnerable to unreliable modalities, where degraded observations may interfere with reliable modality-specific evidence and lead to suboptimal predictions. Moreover, modality reliability can vary across both global driving scenes and individual object queries, requiring adaptive fusion decisions at a finer granularity. To bridge this gap, we reformulate robust camera-LiDAR fusion as a scene-aware branch routing problem and propose SARFusion, a robust 3D object detector. Instead of producing detections from a single fused representation, SARFusion decouples object-query decoding into three parallel reasoning branches: a camera branch, a LiDAR branch, and a camera-LiDAR fusion branch. Guided by a Scene Reliability Prior estimated from the global driving context, SARFusion further incorporates object-level evidence to route each query to the most suitable branch. This query-wise routing strategy alleviates harmful cross-modal interference while preserving the benefits of multimodal fusion when complementary cues are trustworthy. On the nuScenes test set, SARFusion achieves strong performance with 72.5 mAP and 74.4 NDS. Extensive analyses demonstrate its robustness under challenging conditions, including sensor corruptions and environmental changes.

\end{abstract}

\section{Introduction}
\label{intro}

Reliable 3D object detection is a fundamental capability for autonomous driving systems. Modern autonomous vehicles perceive their surroundings with multiple complementary sensors, among which cameras and LiDAR are two of the most widely used modalities. Cameras provide dense appearance and semantic information, which is beneficial for object recognition and distant-object perception. LiDAR directly measures depth and 3D geometry, providing accurate localization cues. By combining these complementary signals, camera-LiDAR fusion detectors have achieved strong performance on standard 3D detection benchmarks.
~\cite{transfusion, deepinteraction, chen2023futr3dunifiedsensorfusion}

Despite this progress, robust multi-modal perception remains challenging in complex driving environments. Sensor observations can be degraded by adverse weather~\cite{multicorrupt}, illumination changes, motion blur, temporal inconsistency, spatial misalignment, camera failures, LiDAR beam dropping, incomplete echoes, and other corruptions~\cite{bijelic2020seeingfogseeingfog,Pitropov_2020,hahner2021fogsimulationreallidar,albreiki2022robustness3dobjectdetectors,multicorrupt,nuscene_c}. Under such conditions, the reliability of each modality can change significantly. A modality that provides useful evidence in clean scenes may become noisy or misleading when degraded. If a detector always follows the same fusion pathway, unreliable modality information can be injected into the final prediction, leading to negative transfer and reduced robustness.

A common limitation of existing fusion detectors is that they treat fusion behavior as largely fixed~\cite{transfusion, bevfusion}. This design implicitly assumes that the same sensor combination is suitable for all driving scenes. However, modality reliability is strongly scene-dependent. Weather, illumination, and sensor status can change the overall usefulness of camera and LiDAR cues. For example, low-light or adverse weather may weaken visual observations, while LiDAR corruptions can reduce geometric reliability. This motivates the need for scene-aware reliability modeling rather than condition-agnostic fusion.

Moreover, modality reliability is not only scene-dependent but also object-dependent. In the same frame, different objects may have different observation quality due to distance, occlusion, visibility, and point density~\cite{nuscene_c,robo3d,robobev}. A distant pedestrian may have sparse LiDAR returns but still preserve recognizable image semantics, while a nearby vehicle may be better localized by LiDAR geometry. Therefore, a single scene-level fusion decision is insufficient: forcing all objects in a frame to use the same sensor combination may still propagate locally degraded evidence. Since object queries naturally represent candidate instances in transformer-based 3D detectors, query-level branch selection provides an appropriate granularity for reliable fusion~\cite{wang2021detr3d,liu2022petr,transfusion,cmt}..

Motivated by these observations, we propose SARFusion, a scene-aware branch routing framework for robust camera-LiDAR 3D object detection. SARFusion explicitly maintains three candidate branches: a camera branch, a LiDAR branch, and a camera-LiDAR fusion branch. These branches provide different sensing paths for object queries under varying sensor conditions. To guide branch selection, SARFusion first estimates a Scene Reliability Prior from global driving conditions such as weather and time. During training, the scene prior is aligned with textual scene prompts~\cite{radford2021learningtransferablevisualmodels}. It then performs Scene-Aware Branch Routing, which combines the scene prior with local feature evidence around each object query to select a reliable branch. This design allows different objects in the same scene to rely on different sensor combinations, reducing negative transfer from degraded modalities while preserving useful cross-modal complementarity.
We evaluate SARFusion on clean and corrupted autonomous-driving benchmarks. In nuScenes~\cite{nuScenes}, SARFusion achieves 71.1 mAP and 73.7 NDS on the validation split, and 72.5 mAP and 74.4 NDS on the test split. Under adverse nuScenes-C conditions~\cite{nuscene_c,huang2024improvingrobustnesslidarcamerafusion}, SARFusion improves detection performance under adverse weather and illumination changes, including fog, snow, rain and strong sunlight. Ablation studies verify the effectiveness of our core components.

Our main contributions are summarized as follows:

\begin{enumerate}
   \item We propose {\it SARFusion}, a scene-aware branch routing framework for robust camera-LiDAR 3D object detection, which dynamically selects among camera, LiDAR, and camera-LiDAR fusion branches instead of relying on a fixed fusion pathway.
   \item We introduce a Scene Reliability Prior to model scene-dependent modality reliability from global driving conditions, providing condition-aware guidance under degraded environments. 
   \item We design Scene-Aware Branch Routing, which combines the scene reliability prior with local object-level evidence to perform query-level branch selection. 
    \item SARFusion achieved SOTA performance among existing camera-LiDAR fusion methods under various sensor failure conditions and extreme weather scenarios.
   \item The code will be publicly available. 
\end{enumerate}

\section{Related Work}
\label{sec:related}

\subsection{3D Object Detection}
3D object detection estimates object categories and 3D bounding boxes from onboard sensor observations. Camera-based methods infer 3D layouts from monocular, stereo, or surround-view images. Early methods rely on geometric priors, keypoints, or explicit depth estimation~\cite{chen2016monocular,brazil2019m3d,liu2020smoke,wang2021fcos3dfullyconvolutionalonestage}. Recent surround-view detectors lift image features into 3D or bird's-eye-view (BEV) space~\cite{huang2022bevdet,li2022bevformer,li2022bevdepth}, or use 3D queries to aggregate multi-view image features~\cite{wang2021detr3d,liu2022petr}. These methods benefit from dense appearance cues, but remain limited by depth ambiguity, especially for distant, occluded, or weakly textured objects.

LiDAR-based detectors exploit point clouds with accurate depth and geometry. Point-based methods operate on raw point sets~\cite{qi2017pointnet,shi2019pointrcnn}, while voxel- and pillar-based methods convert point clouds into regular representations for efficient feature extraction~\cite{zhou2018voxelnet,yan2018second,lang2019pointpillars,yin2021center}. Hybrid point-voxel methods further balance geometric precision and computational efficiency~\cite{shi2021pvrcnn}. However, LiDAR points are sparse at long range and provide limited semantic appearance. They can also be degraded by adverse weather, beam sparsity, incomplete echoes, and sensor malfunction. These complementary limitations motivate camera-LiDAR fusion for more accurate and reliable 3D detection.


\subsection{Multi-modal Sensor Fusion}
Camera-LiDAR fusion combines dense image semantics with accurate LiDAR geometry. Point-level methods project image semantics or features onto LiDAR points to enrich point-cloud representations~\cite{pointpainting,pointaugmenting,mvp}. They improve detection performance, but usually depend on accurate calibration and are sensitive to projection errors, spatial misalignment, and LiDAR degradation.

Recent methods fuse modalities in intermediate feature spaces. BEV-based methods transform camera and LiDAR features into a shared BEV representation~\cite{bevfusion,unitr}, which provides an efficient fusion space but still relies on depth estimation and geometric projection. Other methods explore unified voxel representations, deformable feature alignment, and sparse multi-sensor representations~\cite{li2022unifyingvoxelbasedrepresentationtransformer,chen2023futr3dunifiedsensorfusion,chen2022autoalignv2deformablefeatureaggregation,xie2023sparsefusionfusingmultimodalsparse}. Transformer-based methods use object queries or modality tokens for cross-modal interaction~\cite{transfusion,deepinteraction,cmt}. TransFusion refines LiDAR proposals with image features, DeepInteraction performs iterative modality interaction, and CMT uses coordinate embeddings for implicit alignment. These methods achieve strong clean-set performance, but most of them use fixed fusion pathways or predefined interaction patterns. They rarely adapt the fusion behavior to scene-dependent and object-dependent modality reliability.


\subsection{Robust Multi-Modality Fusion}
Robust multi-modal fusion has attracted increasing attention because 3D detectors can degrade severely under adverse weather, motion blur, temporal inconsistency, spatial misalignment, missing cameras, beam dropping, and incomplete echoes~\cite{robobev,robo3d,multicorrupt,ji2025enhancing,ji2026ill}. These corruptions change modality reliability at different granularities. Weather and illumination often affect the whole scene, while occlusion, distance, point density, and local sensor failures may affect individual objects within the same frame.

Existing methods improve robustness through data augmentation, modality dropout, decoupled representations, or adaptive fusion strategies~\cite{cmt,metabev,unitr,meformer,mome,drews2022deepfusionrobustmodular3d,huang2024improvingrobustnesslidarcamerafusion,sural2024contextualfusioncontextbasedmultisensorfusion,sadeghian2025reliabilitydrivenlidarcamerafusionrobust}. CMT adopts masked-modal training to handle missing modalities~\cite{cmt}. MEFormer reduces negative fusion with modality-agnostic decoding and prediction ensemble~\cite{meformer}. Recent expert-based methods further explore routing or selecting modality-specific experts under sensor failures~\cite{mome}. These studies suggest that reducing rigid inter-modality dependence is important for robust perception.

However, many existing robust fusion methods mainly focus on predefined corruptions, modality-level failures, or fixed robustness training strategies. Less attention has been paid to jointly modeling global scene reliability and query-level local observation quality. In realistic driving scenes, the reliable modality may vary not only with weather or illumination, but also across different objects due to distance, visibility, occlusion, and point density~\cite{ji2025visualtrans,ji2025mathsticks,tian2026spatial,wang2026orca}. Motivated by this observation, SARFusion formulates robust camera-LiDAR fusion as a scene-aware branch routing problem. It estimates a global scene reliability prior and combines it with local query evidence to select camera, LiDAR, or fusion branches for each object query~\cite{ji2026prm,liu2026prm}.

\section{Method}
\subsection{Overview}

\begin{figure*}[t]
  \centering
  \includegraphics[width=\linewidth]{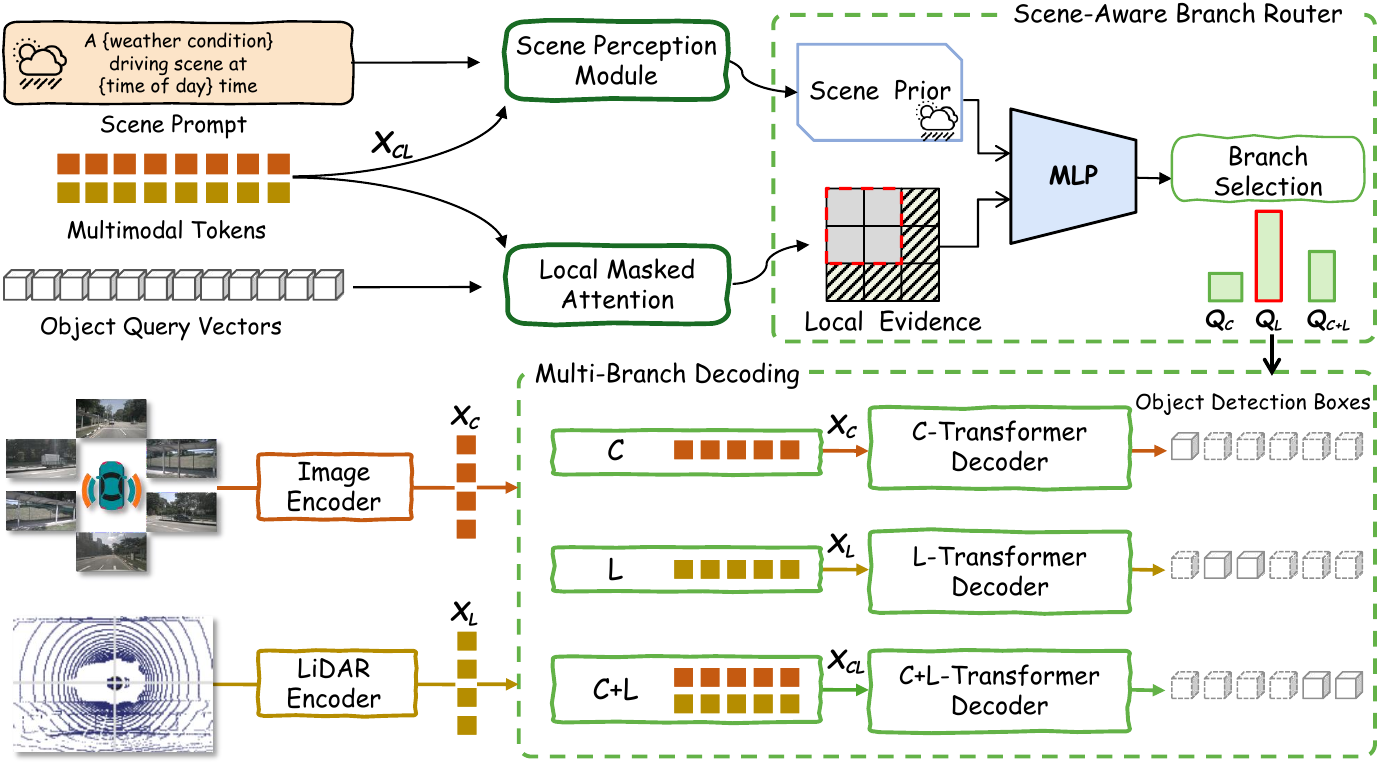}
    \caption{Overall architecture of the proposed SARFusion.}
    \label{fig:SARFusion}
    \vspace{-0.5em}
\end{figure*}

The overall framework of SARFusion is shown in Fig.~\ref{fig:SARFusion}. Given multi-view camera images $\mathcal{I}=\{I_v\}_{v=1}^{V}$ and a LiDAR point cloud $\mathcal{P}$, SARFusion first extracts camera features and LiDAR features using modality-specific encoders. The camera features are flattened into image tokens $\mathbf{X}_C \in \mathbb{R}^{N_C \times D}$, and the LiDAR features are encoded into BEV tokens $\mathbf{X}_L \in \mathbb{R}^{N_L \times D}$. We further concatenate them as multimodal context tokens $\mathbf{X}_{CL}=[\mathbf{X}_C;\mathbf{X}_L]$.

Instead of producing a single fused representation, SARFusion explicitly maintains three candidate decoding branches: a camera branch, a LiDAR branch, and a Camera-LiDAR fusion branch. These branches provide complementary representation choices under different sensing conditions. The camera branch decodes object queries using camera tokens, the LiDAR branch uses LiDAR tokens, and the fusion branch attends to multimodal tokens.
To adaptively select reliable branches, SARFusion first learns a Scene Reliability Prior from multimodal context to encode global sensing conditions. Then, for each object query, a Scene-Aware Branch Router combines this scene prior with local observation evidence around the query reference point and predicts branch probabilities over the three decoding branches. The query is routed to the selected branch for final decoding and prediction.

\subsection{Scene Reliability Prior}

The reliability of camera and LiDAR observations is highly correlated with scene-level conditions, such as weather, illumination, sensor degradation, and temporal-spatial disturbance. Therefore, we introduce a Scene Reliability Prior to provide global reliability guidance for branch routing.

Given multimodal context tokens $\mathbf{X}_{CL}$, we use a lightweight Transformer to aggregate scene-level information. Specifically, a learnable scene token $\mathbf{q}_s$ is prepended to the multimodal tokens, and the output scene token is projected to obtain the scene reliability prior:
\begin{equation}
    \mathbf{c}_{prior} = \mathrm{LN}\left(\mathrm{MLP}\left(
    \mathrm{Transformer}([\mathbf{q}_s;\mathbf{X}_{CL}])_0
    \right)\right),
\end{equation}
where $\mathbf{c}_{prior} \in \mathbb{R}^{D}$ summarizes the global sensing condition of the current scene.

To encourage the scene prior to encode meaningful reliability-related information, we supervise it with textual scene prompts during training. For each training scene, a prompt describing the scene condition is constructed, e.g.,
\begin{equation}
    \text{``A \{weather condition\} driving scene at \{time of day\}.''}
\end{equation}
The prompt is encoded by a text encoder into a text embedding $\mathbf{t}$. We then apply a symmetric contrastive loss between the scene prior and the text embedding:
\begin{equation}
    \mathcal{L}_{srp}
    =
    \mathcal{L}_{c \rightarrow t}
    +
    \mathcal{L}_{t \rightarrow c}.
\end{equation}
This supervision encourages $\mathbf{c}_{prior}$ to preserve scene-level reliability cues. During inference, no text prompt or scene label is required; the scene prior is directly inferred from multimodal features.

\begin{figure*}[t]
  \centering
  \includegraphics[width=1\linewidth]{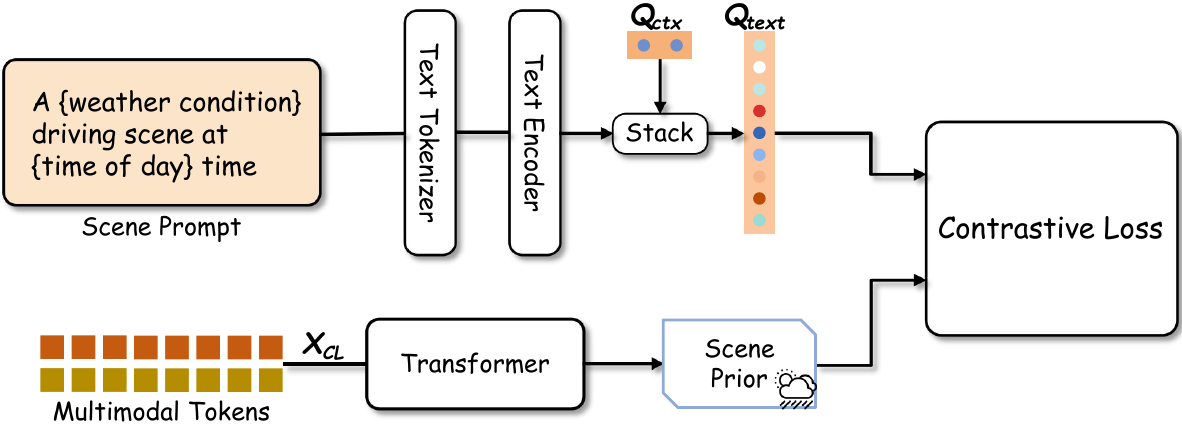}
    \caption{The illustration of SRP module}
    \label{fig:SRP}
    \vspace{-0.5em}
\end{figure*}

\subsection{Scene-Aware Branch Routing}
While the Scene Reliability Prior captures global sensing conditions, modality reliability can also vary across object regions~\cite{nuscene_c,robo3d,robobev}. For example, LiDAR points may be sparse for distant objects, while camera observations may be locally occluded or corrupted. Therefore, SARFusion performs branch routing at the object-query level.

For the $i$-th object query $\mathbf{q}_i$ with reference point $\mathbf{p}_{i}^{ref}=(x_i,y_i,z_i)$, we project the reference point onto the camera feature plane and the LiDAR BEV plane. Around each projected location, we construct a local attention window to collect nearby modality features. The union of camera and LiDAR local tokens forms a query-specific local region $\Omega_i$.

We use a local attention mask $\mathbf{M}_i$ to restrict attention to $\Omega_i$:
\begin{equation}
    \mathbf{M}_i(n)=
    \begin{cases}
        0, & n \in \Omega_i, \\
        -\infty, & n \notin \Omega_i.
    \end{cases}
\end{equation}
The local observation evidence for query $\mathbf{q}_i$ is then computed by masked attention:
\begin{equation}
    \mathbf{e}^{local}_i
    =
    \mathrm{Softmax}
    \left(
    \frac{\mathbf{q}_i \mathbf{K}^{\top}}{\sqrt{D}} + \mathbf{M}_i
    \right)
    \mathbf{V},
\end{equation}
where $\mathbf{K}$ and $\mathbf{V}$ are projected from multimodal context tokens $\mathbf{X}_{CL}$.

The branch router combines local observation evidence with the global scene prior:
\begin{equation}
    \mathbf{h}_i = [\mathbf{e}^{local}_i;\mathbf{c}_{prior}],
\end{equation}
and predicts branch probabilities:
\begin{equation}
    \boldsymbol{\pi}_i
    =
    \mathrm{Softmax}
    \left(
    \mathrm{MLP}(\mathbf{h}_i)
    \right),
    \quad
    \boldsymbol{\pi}_i =
    [\pi_{i,C}, \pi_{i,L}, \pi_{i,F}],
\end{equation}
where $C$, $L$, and $F$ denote the camera, LiDAR, and fusion branches, respectively. The selected branch is:
\begin{equation}
    m_i^{*} = \arg\max_{m \in \{C,L,F\}} \pi_{i,m}.
\end{equation}

Accordingly, object queries are partitioned into three groups:
\begin{equation}
    \mathcal{Q}_{m}
    =
    \{\mathbf{q}_i \mid m_i^{*}=m\},
    \quad
    m \in \{C,L,F\}.
\end{equation}
Each group is decoded by its corresponding branch. This design allows different object queries in the same scene to select different reliable representations.

\begin{figure*}[t]
  \centering
  \includegraphics[width=1\linewidth]{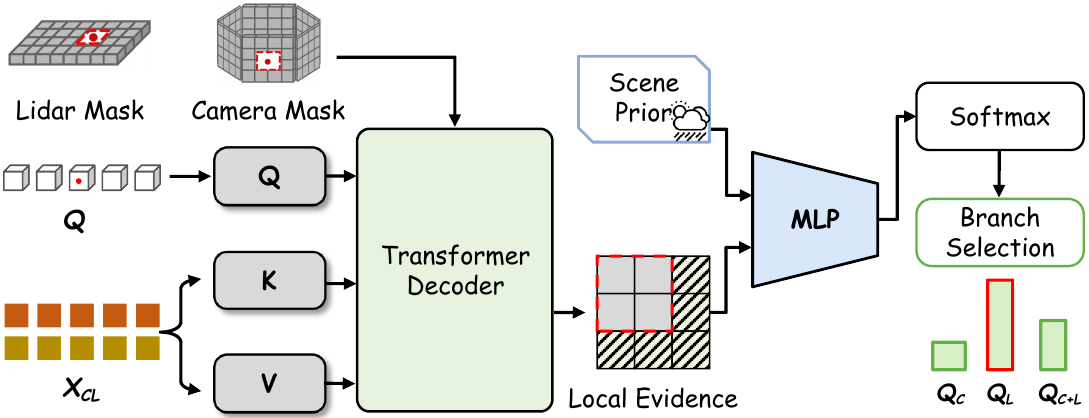}
    \caption{The illustration of SABR module.}
    \label{fig:DER}
    \vspace{-0.5em}
\end{figure*}

\subsection{Training Objective}

Training SARFusion involves three objectives: branch-wise detection supervision, scene prior supervision, and routing supervision.

First, to ensure that each branch has independent detection ability, we apply detection losses to all three branches:
\begin{equation}
    \mathcal{L}_{det}
    =
    \mathcal{L}_{det}^{C}
    +
    \mathcal{L}_{det}^{L}
    +
    \mathcal{L}_{det}^{F},
\end{equation}
where each detection loss consists of classification and box regression terms.

Second, the Scene Reliability Prior is supervised by the contrastive scene-prompt loss $\mathcal{L}_{srp}$ defined above, which encourages the prior to encode global sensing conditions.

Third, we supervise the branch router with reliability-aware routing targets. During training, modality degradation or dropout is applied to simulate unreliable sensing conditions. The routing target encourages queries to select the branch corresponding to the reliable modality: the LiDAR branch when camera observations are degraded, the camera branch when LiDAR observations are unreliable, and the fusion branch when both modalities are informative. The routing loss is:
\begin{equation}
    \mathcal{L}_{route}
    =
    -\frac{1}{N}
    \sum_{i=1}^{N}
    \sum_{m \in \{C,L,F\}}
    y_{i,m}\log \pi_{i,m},
\end{equation}
where $y_{i,m}$ is the routing target for query $\mathbf{q}_i$.

The full training objective is:
\begin{equation}
    \mathcal{L}
    =
    \lambda_{det}\mathcal{L}_{det}
    +
    \lambda_{srp}\mathcal{L}_{srp}
    +
    \lambda_{route}\mathcal{L}_{route}.
\end{equation}

In practice, we adopt a staged training strategy. We first train the three decoding branches with branch-wise detection supervision, then learn the Scene Reliability Prior with prompt-based contrastive supervision, and finally optimize the router with reliability-aware routing supervision. This strategy stabilizes branch learning and prevents the router from degenerating into a fixed branch preference.

\section{Experiments}
\subsection{Experimental Setup}

\textbf{Dataset and metrics.}
We evaluate SARFusion on the nuScenes 3D object detection benchmark, which provides synchronized multi-view camera images and LiDAR point clouds for autonomous driving scenes. Following the official protocol, the dataset is divided into 700 training scenes, 150 validation scenes, and 150 test scenes. We report the official mean Average Precision (mAP) and nuScenes Detection Score (NDS). mAP measures localization and classification accuracy, while NDS further summarizes multiple detection quality factors, including translation, scale, orientation, velocity, and attribute errors.

\textbf{Robustness evaluation.}
Besides the standard clean validation and test sets, we further evaluate different methods under representative degraded sensing conditions, including snow, rain, fog, and strong sunlight. These conditions affect camera and LiDAR observations in different ways. Snow, rain, and fog may weaken image visibility and disturb point-cloud measurements, while strong sunlight mainly changes image appearance and reduces the reliability of visual cues. Unless otherwise specified, all models are trained on the original training set and directly evaluated under degraded validation conditions without condition-specific fine-tuning. This protocol is used to examine whether the fusion strategy itself can adapt to scene-dependent sensing reliability.

\begin{figure*}[t]
  \centering
  \includegraphics[width=1\linewidth]{\detokenize{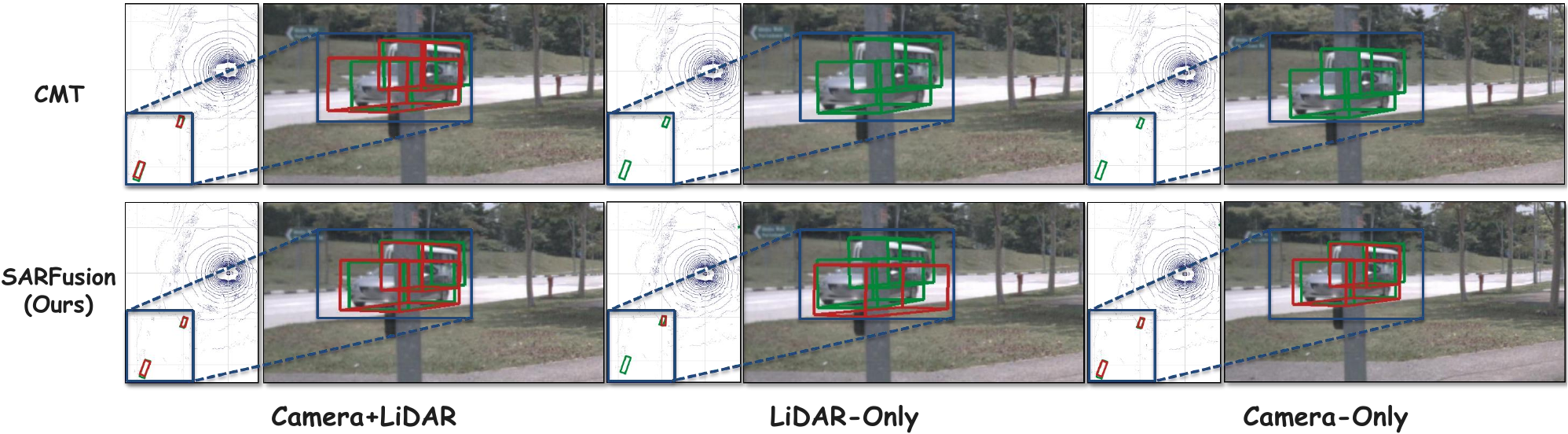}}
  \caption{Comparison of camera-LiDAR observations under degraded sensing conditions.}
  \label{fig:Camera_Lidar_CMP}
  \vspace{-0.5em}
\end{figure*}

\textbf{Implementation details.}
SARFusion follows a multi-branch camera-LiDAR detection framework. The camera stream extracts image features from surrounding-view images, and the LiDAR stream encodes point clouds into BEV features. Based on these representations, SARFusion maintains three candidate prediction branches: a camera branch, a LiDAR branch, and a camera-LiDAR fusion branch. The Scene Reliability Prior estimates global scene-level sensing conditions, while the Scene-Aware Branch Routing module combines this prior with local query evidence to select a suitable branch for each object query. All ablation experiments are conducted on the nuScenes validation split. We adopt a staged training strategy: the detection branches are first optimized to provide valid modality-specific and fused representations; the scene prior is then learned with scene-level supervision; finally, the routing module is jointly optimized with the detection objective.

\subsection{Comparison with SOTA Methods}

\textbf{Results on the clean benchmark.}
Table~\ref{tab:clean_comparison} compares SARFusion with representative camera-only, LiDAR-only, and camera-LiDAR fusion methods on the standard nuScenes validation and test sets. Camera-only methods are limited by the lack of explicit 3D geometry, while LiDAR-only methods provide stronger localization due to accurate depth and structure. Multi-modal methods further improve detection by combining image semantics with LiDAR geometry.

SARFusion achieves 71.1 mAP and 73.7 NDS on the validation set, and 72.5 mAP and 74.4 NDS on the test set. Compared with representative camera-LiDAR fusion methods, SARFusion obtains strong performance on clean scenes. The improvement on clean data is moderate but meaningful, since most clean samples already benefit from conventional fused representations. This indicates that the proposed scene-aware routing strategy improves robustness without sacrificing standard detection accuracy.

\begin{table*}[t]
\centering
\caption{Comparison with representative 3D object detection methods on the nuScenes validation and test sets. All results are reported in percentage.}
\label{tab:clean_comparison}
\setlength{\tabcolsep}{3.5pt}
\resizebox{0.9\textwidth}{!}{
\footnotesize
\begin{tabular}{llcccc}
\toprule
\multirow{2}{*}{Modality} & \multirow{2}{*}{Method} & \multicolumn{2}{c}{Validation} & \multicolumn{2}{c}{Test} \\
\cmidrule(lr){3-4} \cmidrule(lr){5-6}
 & & mAP & NDS & mAP & NDS \\
\midrule
\multirow{3}{*}{Camera}
& FCOS3D~\cite{wang2021fcos3dfullyconvolutionalonestage} & 34.3 & 41.5 & 35.8 & 42.8 \\
& PETR~\cite{liu2022petr} & 37.0 & 44.2 & 39.1 & 45.5 \\
& BEVDet~\cite{huang2022bevdet} & -- & -- & 42.2 & 48.2 \\
\midrule
\multirow{3}{*}{LiDAR}
& SECOND~\cite{yan2018second} & 52.6 & 63.0 & 52.8 & 63.3 \\
& CenterPoint~\cite{yin2021center} & 59.6 & 66.8 & 60.3 & 67.3 \\
& TransFusion-L~\cite{transfusion} & 65.1 & 70.1 & 65.5 & 70.2 \\
\midrule
\multirow{10}{*}{Camera + LiDAR}
& FUTR3D~\cite{chen2023futr3dunifiedsensorfusion} & 64.5 & 68.3 & -- & -- \\
& PointAugmenting~\cite{pointaugmenting} & -- & -- & 66.8 & 71.0 \\
& UVTR~\cite{li2022unifyingvoxelbasedrepresentationtransformer} & 65.4 & 70.2 & 67.1 & 71.1 \\
& AutoAlignV2~\cite{chen2022autoalignv2deformablefeatureaggregation} & 67.1 & 71.2 & 68.4 & 72.4 \\
& TransFusion~\cite{transfusion} & 67.5 & 71.3 & 68.9 & 71.6 \\
& MetaBEV~\cite{metabev} & 68.0 & 71.5 & -- & -- \\
& BEVFusion~\cite{bevfusion} & 68.5 & 71.4 & 70.2 & 72.9 \\
& DeepInteraction~\cite{deepinteraction} & 69.9 & 72.6 & 70.8 & 73.4 \\
& SparseFusion~\cite{xie2023sparsefusionfusingmultimodalsparse} & 70.4 & 72.8 & 72.0 & 73.8 \\
& CMT~\cite{cmt} & 70.3 & 72.9 & 72.0 & 74.1 \\
\midrule
Camera + LiDAR
& \textbf{SARFusion} & \textbf{71.1} & \textbf{73.7} & \textbf{72.5} & \textbf{74.4} \\
\bottomrule
\end{tabular}}
\end{table*}

\textbf{Results under degraded conditions.}
Table~\ref{tab:robust_comparison} reports the robustness comparison under representative weather and lighting conditions. Compared with clean scenes, degraded scenes introduce stronger modality imbalance. Visual appearance may become unreliable under rain, fog, or strong sunlight, while LiDAR observations can become sparse or noisy in adverse weather. These corruptions make tightly coupled fusion vulnerable, since degraded observations may interfere with reliable modality-specific evidence.

SARFusion achieves the best performance under all evaluated clean and degraded conditions. Compared with the strongest competing results, SARFusion improves mAP by 0.8 on clean scenes, 1.3 under snow, 3.2 under rain, 7.8 under fog, and 2.0 under strong sunlight. The advantage becomes more evident under degraded conditions, supporting our motivation that robust camera-LiDAR fusion should adapt to scene-level and object-level reliability rather than relying on a single fixed fused representation.

\begin{figure*}[t]
  \centering
  \includegraphics[width=0.8\linewidth]{\detokenize{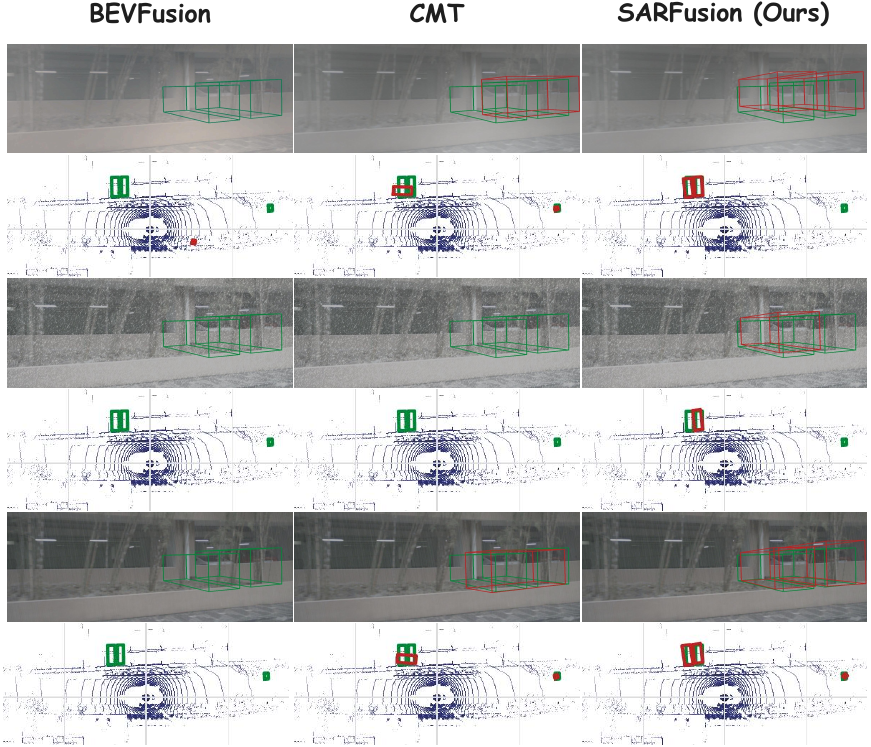}}
  \caption{Qualitative detection results under representative adverse weather conditions, including fog, snow, and rain.}
  \label{fig:Snow_Fog_Rain}
  \vspace{-0.5em}
\end{figure*}

\begin{table*}[t]
\centering
\caption{Robustness comparison on the nuScenes validation set under representative weather and lighting conditions.}
\label{tab:robust_comparison}
\setlength{\tabcolsep}{5.0pt}
\resizebox{\textwidth}{!}{
\begin{tabular}{lccccc}
\toprule
Method & Clean & Snow & Rain & Fog & Strong Sunlight \\
\midrule
FCOS3D~\cite{wang2021fcos3dfullyconvolutionalonestage} & 23.9 & 2.0 & 13.0 & 13.5 & 17.2 \\
DETR3D~\cite{wang2021detr3d} & 34.7 & 5.1 & 20.4 & 27.9 & 34.7 \\
PointPillars~\cite{lang2019pointpillars} & 27.7 & 27.6 & 27.7 & 24.5 & 23.7 \\
CenterPoint~\cite{yin2021center} & 59.3 & 55.9 & 56.1 & 43.8 & 54.2 \\
BEVFusion~\cite{bevfusion} & 68.5 & 62.8 & 66.1 & 54.1 & 64.4 \\
TransFusion~\cite{transfusion} & 66.4 & 63.3 & 65.4 & 53.7 & 55.1 \\
DeepInteraction~\cite{deepinteraction} & 69.9 & 62.3 & 66.5 & 54.8 & 64.9 \\
FUTR3D~\cite{chen2023futr3dunifiedsensorfusion} & 64.2 & 52.7 & 58.4 & 53.2 & 57.7 \\
CMT~\cite{cmt} & 70.3 & 63.5 & 62.6 & 61.4 & 66.3 \\
\midrule
\textbf{SARFusion} & \textbf{71.1} & \textbf{64.8} & \textbf{69.7} & \textbf{69.2} & \textbf{68.3} \\
\bottomrule
\end{tabular}}
\end{table*}

\subsection{Ablation Study}

\textbf{Effectiveness of each component.}
Table~\ref{tab:component_ablation} studies the contribution of the main components in SARFusion. The single fusion branch baseline already achieves strong performance, since it benefits from both camera semantics and LiDAR geometry. However, directly introducing multiple branches without an effective routing mechanism leads to a significant performance drop. This shows that simply adding modality-specific branches is not sufficient. Without query-level routing, the detector cannot determine which representation should be responsible for each object, and the branch predictions are not properly coordinated.

After introducing Scene-Aware Branch Routing, the performance recovers to the level of the single fusion branch baseline. This result indicates that query-level branch selection is essential for making the multi-branch design effective. Finally, adding the Scene Reliability Prior further improves both mAP and NDS, showing that global scene reliability provides useful context for object-level routing decisions.

\begin{table}[t]
\centering
\caption{Component ablation of SARFusion on the nuScenes validation set.}
\label{tab:component_ablation}
\setlength{\tabcolsep}{5.5pt}
\resizebox{0.7\columnwidth}{!}{
\begin{tabular}{lcc}
\toprule
Setting & mAP & NDS \\
\midrule
Single fusion branch & 70.7 & 73.1 \\
Multi-branch w/o routing & 52.4 & 62.8 \\
+ Scene-Aware Branch Routing & 70.8 & 73.1 \\
+ Scene Reliability Prior & \textbf{71.1} & \textbf{73.7} \\
\bottomrule
\end{tabular}}
\end{table}

\textbf{Effect of scene prior supervision.}
Table~\ref{tab:scene_prior_ablation} evaluates the effect of supervising the Scene Reliability Prior. Without scene-level supervision, the prior is learned only through the final detection objective. In this case, the scene representation may encode information useful for detection, but it is not explicitly encouraged to describe global sensing reliability.

With scene prior supervision, SARFusion achieves higher mAP and NDS. This improvement shows that explicitly aligning the scene prior with global driving conditions makes the routing decision more stable. The prior does not replace local evidence; instead, it calibrates local query evidence with scene-level reliability context.

\begin{table}[t]
\centering
\caption{Effect of scene prior supervision on the nuScenes validation set.}
\label{tab:scene_prior_ablation}
\setlength{\tabcolsep}{6pt}
\begin{tabular}{lcc}
\toprule
Setting & mAP & NDS \\
\midrule
w/o scene prior supervision & 69.6 & 72.1 \\
w/ scene prior supervision & \textbf{71.1} & \textbf{73.7} \\
\bottomrule
\end{tabular}
\end{table}

\textbf{Effect of routing evidence.}
Table~\ref{tab:routing_evidence_ablation} compares different inputs for branch routing. Using only local query evidence already provides strong performance, because it captures object-level cues such as local image visibility, point density, and feature consistency around the query reference point. This is important for autonomous driving scenes, where degradation is often spatially non-uniform.

Combining local evidence with the Scene Reliability Prior gives the best result. The two types of evidence are complementary: local evidence describes the reliability of a specific object query, while the scene prior describes the global sensing condition shared by the whole frame. Their combination enables SARFusion to make object-specific routing decisions under a scene-aware context.

\begin{table}[t]
\centering
\caption{Ablation of routing evidence on the nuScenes validation set.}
\label{tab:routing_evidence_ablation}
\setlength{\tabcolsep}{6pt}
\begin{tabular}{lcc}
\toprule
Routing Evidence & mAP & NDS \\
\midrule
Local evidence only & 70.8 & 73.1 \\
Scene prior + local evidence & \textbf{71.1} & \textbf{73.7} \\
\bottomrule
\vspace{-0.7em}
\end{tabular}
\end{table}

\textbf{Effect of training strategy.}
Table~\ref{tab:training_strategy_ablation} compares direct end-to-end training with the staged training strategy. Direct training gives much lower performance, indicating that joint optimization of branch specialization, scene prior learning, and routing selection is difficult from scratch. In the early training stage, the router may make unstable choices before each branch becomes a reliable detector, which further weakens branch learning and may lead to routing collapse.

The staged strategy substantially improves performance. By first training the candidate branches, each branch learns a meaningful representation for detection. The scene prior is then learned as a reliability-aware condition signal. Finally, the routing module is optimized on top of already informative branches and a stable scene prior.

\begin{table}[t]
\centering
\caption{Effect of training strategy on the nuScenes validation set.}
\label{tab:training_strategy_ablation}
\setlength{\tabcolsep}{5pt}
\begin{tabular}{lcc}
\toprule
Training Strategy & mAP & NDS \\
\midrule
Direct training & 65.7 & 69.6 \\
Staged training & \textbf{71.1} & \textbf{73.7} \\
\bottomrule
\vspace{-0.7em}
\end{tabular}
\end{table}

\subsection{Qualitative Analysis}

\begin{figure*}[t]
  \centering
  \includegraphics[width=0.9\linewidth]{\detokenize{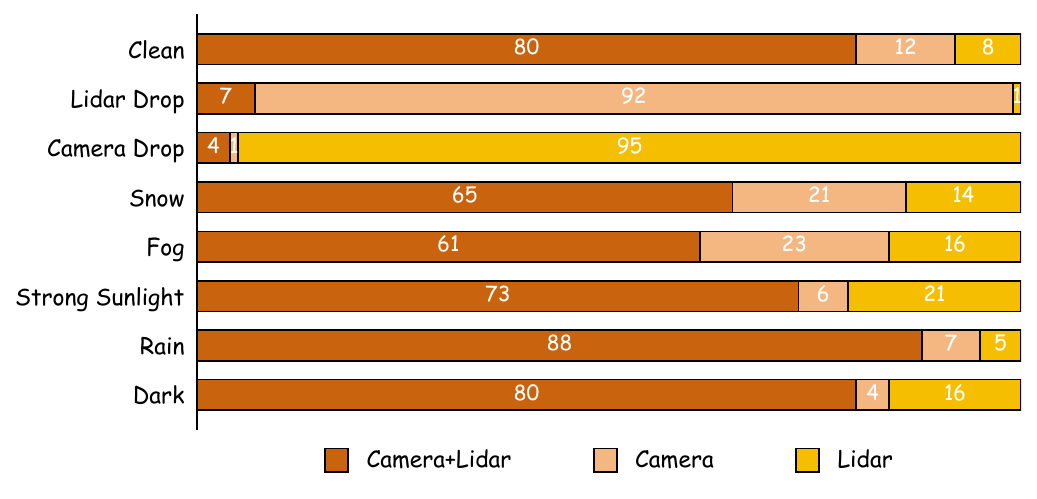}}
  \caption{Branch routing visualization under different sensing conditions.}
  \label{fig:branch_routing_visualization}
  \vspace{-1em}
\end{figure*}

\textbf{Routing behavior under different scenes.}
To further understand how SARFusion makes routing decisions, we analyze the distribution of selected branches under different sensing conditions. As shown in Figure~\ref{fig:branch_routing_visualization}, the fusion branch dominates in clean scenes, which is consistent with the strong performance of conventional fusion methods under normal conditions. This confirms that SARFusion does not unnecessarily avoid fusion when both modalities are reliable.

When one modality becomes unreliable, the routing distribution changes accordingly. Under LiDAR failure, most queries are routed to the camera branch, while under camera failure, most queries are routed to the LiDAR branch. Under adverse weather or illumination, the model does not simply switch to a single modality. Instead, it keeps a considerable portion of queries in the fusion branch while increasing the use of modality-specific branches. This behavior indicates that SARFusion learns a reliability-aware routing policy rather than a fixed modality preference.

\textbf{Detection examples.}
Figure~\ref{fig:Snow_Fog_Rain} presents qualitative comparisons under representative adverse weather conditions, including rain, fog, and snow. These conditions degrade image visibility and may disturb point-cloud observations, making tightly coupled camera-LiDAR fusion vulnerable to unreliable modality features. Compared with BEVFusion and CMT, SARFusion produces more complete and accurately localized detection results across the degraded scenes.

In particular, competing fusion methods may miss objects or generate less precise boxes when corrupted observations interfere with reliable modality-specific evidence. By routing object queries to camera, LiDAR, or fusion branches according to scene-level and object-level reliability, SARFusion alleviates harmful cross-modal interference while preserving useful complementary cues. These qualitative results further support the robustness of scene-aware routing fusion under adverse weather conditions.

\section{Conclusions}

We introduced \textit{SARFusion}, a scene-aware branch routing framework for robust camera-LiDAR 3D object detection. Unlike fixed fusion methods, SARFusion maintains camera, LiDAR, and fusion branches, and combines a global Scene Reliability Prior with local query evidence to select reliable sensing paths for each object. This design reduces the influence of degraded modalities while preserving complementary semantic and geometric cues. Experiments on nuScenes and corrupted driving scenarios demonstrate the effectiveness of SARFusion: it achieves 72.5 mAP and 74.4 NDS on the nuScenes test set and consistently improves robustness under snow, rain, fog, and strong sunlight. Ablations further validate the benefits of scene reliability modeling and query-level routing.

%
%
\bibliographystyle{splncs04}
\bibliography{main}
\end{document}